\documentclass{article}

\usepackage[preprint]{neurips_2026}

\usepackage[utf8]{inputenc}
\usepackage[T1]{fontenc}
\usepackage{hyperref}
\usepackage{url}
\usepackage{microtype}
\usepackage{graphicx}
\usepackage{subcaption}
\usepackage{booktabs}
\usepackage{makecell}
\usepackage{nicefrac}
\usepackage{xcolor}
\usepackage{amsmath}
\usepackage{amssymb}
\usepackage{amsfonts}
\usepackage{mathtools}
\usepackage{amsthm}
\usepackage{algorithm}
\usepackage{algorithmic}
\usepackage{multirow}
\usepackage{longtable}
\usepackage[capitalize,noabbrev]{cleveref}
\usepackage[textsize=tiny]{todonotes}
\usepackage{pdflscape}
\usepackage{array}
\usepackage{needspace}
\theoremstyle{plain}
\newtheorem{theorem}{Theorem}[section]

\theoremstyle{definition}
\newtheorem{definition}[theorem]{Definition}

\theoremstyle{remark}
\newtheorem{remark}[theorem]{Remark}

\title{Budget-aware Auto Optimizer Configurator}

\author{%
  Kang Liu \\
  School of Future Technology \\
  Xi'an Jiaotong University \\
  Xi'an, China \\
  \texttt{kanyo@foxmail.com}
  \And
  Wei Peng \\
  School of Automation Science and Engineering \\
  Xi'an Jiaotong University \\
  Xi'an, China \\
  \texttt{weipeng@stu.xjtu.edu.cn}
  \And
  Jianchen Hu$^*$\\
  School of Future Technology \\
  School of Automation Science and Engineering \\
  Xi'an Jiaotong University \\
  Xi'an, China \\
  \texttt{horace89@gmail.com}
}

\begin{document}

\maketitle

\begin{abstract} 
Optimizer states occupy massive GPU memory in large-scale model training. However, gradients in different network blocks exhibit distinct behaviors, such as varying directional stability and scale anisotropy, implying that expensive optimizer states are not universally necessary and using a global optimizer is often memory-inefficient. We propose the Budget-Aware Optimizer Configurator (BAOC) to reduce memory cost by assigning suitable optimizer configurations to individual blocks under given budgets. Specifically, BAOC samples gradient streams to derive statistical metrics that quantify the potential performance risk of applying cheaper configurations (e.g., low precision or removing momentum). It then solves a constrained allocation problem to minimize total risk under memory and time budgets, selecting a budget-feasible configuration for each block. Experiments across vision, language, and diffusion workloads demonstrate that BAOC maintains training quality while significantly reducing the memory usage of optimizer states. The code is available at \url{https://anonymous.4open.science/r/BAOC-45C6}.
\end{abstract}

\section{Introduction}

In large-scale model training, optimizer states consume massive GPU memory, as they maintain high-precision running statistics (e.g., first and second moments). This heavy memory cost reduces the available memory for activations, forcing the training process to choose smaller batch sizes, which leads to reduced training throughput and increased gradient noise, hindering efficient convergence.

In order to reduce the memory cost of optimizer states, existing work primarily focuses on optimizing update mechanisms and state precision (we refer to a specific combination of mechanism and precision as a configuration; see Definition~\ref{def:config}). Strategies include low-bit quantization (e.g., 8-bit optimizers \citep{DBLP:conf/iclr/DettmersLSZ22}), factorized statistics (e.g., Adafactor \citep{shazeer2018adafactor}), and low-rank gradient projections (e.g., GaLore \citep{pmlr-v235-zhao24s}). The growing diversity of optimizer designs, each emphasizing different aspects of gradient dynamics, makes configuration selection increasingly nontrivial. However, existing methods typically apply a uniform configuration across the entire network, overlooking block-wise heterogeneity in directional stability, scale anisotropy, and noise levels. This can lead to mechanism--block mismatch: momentum may be useful for stable gradient directions but less effective under strong fluctuations, while adaptive scaling is essential for anisotropic blocks but redundant for nearly uniform ones.

Therefore, a natural solution is block-wise optimizer configuration: assigning different configurations to different blocks. Under a fixed budget, this has two benefits. First, it saves memory by keeping expensive states only where they are needed. Second, it enhances training efficiency by aligning optimizer mechanisms with specific gradient behaviors. Prior studies on layer-wise learning rates \citep{You2018LARS} validate that applying heterogeneous update mechanisms does not inherently destabilize training. Thus, the core challenge lies in systematic decision-making. We must address two key problems: (1) how to quantify the mismatch between a block and a candidate configuration (we refer to this as mismatch risk), and (2) how to select a suitable configuration for each block under given memory and per-step time budgets.

To address these challenges, we propose the Budget-Aware Optimizer Configurator (BAOC). BAOC is an automated framework designed to assign optimizer configurations to parameter blocks under given budgets. It builds on the insight that different blocks require different update mechanisms and tolerate varying levels of quantization. Specifically, BAOC samples gradient streams to compute lightweight statistics aligned with optimizer mechanisms. It then converts these statistics into risk scores and solves a constrained allocation problem to obtain a budget-feasible configuration for each block. Experiments on vision, language, and diffusion workloads show that BAOC achieves a competitive memory--quality trade-off, maintaining training quality with much lower optimizer-state memory usage. The main contributions are as follows:

(1) We propose BAOC, which formulates block-wise optimizer configuration as a constrained resource allocation problem. Unlike global selection, BAOC assigns tailored mechanisms and precision to each parameter block by minimizing the gradient-optimizer mismatch risk under given budgets.

(2) We develop an interpretable, lightweight block-level metrics suite and a practical optimization pipeline. BAOC maps risk signals from sampled gradient statistics (e.g., anisotropy and directional stability) into an optimization problem for budget-aware allocation. Experiments across vision, language, and diffusion workloads demonstrate competitive memory--quality trade-offs.

The remainder of this paper is organized as follows. Section~\ref{sec:related} discusses related work. Section~\ref{sec:method} presents BAOC design. Section~\ref{sec:exp} evaluates BAOC on multi tasks. Section~\ref{sec:discussion} concludes with limitations and future directions.

\section{Related Work}\label{sec:related}

Modern optimizers are specialized for distinct gradient behaviors. Momentum handles high-curvature oscillations \citep{Sutskever2013Momentum}, while adaptive methods (e.g., Adam, RMSProp) address anisotropic gradient scales \citep{Kingma2015Adam}. Recent innovations further refine these mechanisms via decoupled decay \citep{Loshchilov2019AdamW} or sign-based updates \citep{Chen2023Lion}, with some works exploring stateless training \citep{ma2024swan,glentis2025scale}. These diverse designs confirm that no single mechanism is universally optimal for all gradient patterns.

To reduce state memory, research focuses on three paradigms: (1) Quantization stores states in low precision (e.g., 8-bit) without degrading quality \citep{DBLP:conf/iclr/DettmersLSZ22,yao2024shampoo4bit}; (2) Factorization approximates second moments with low-rank statistics \citep{shazeer2018adafactor}; (3) Subspace projection restricts updates to low-rank manifolds to minimize state storage \citep{pmlr-v235-zhao24s,zhang2024qgalore}. Meanwhile, high-performance preconditioners such as Shampoo/SOAP offer stronger conditioning but usually incur higher state cost, whereas stateless designs minimize states but may be less stable \citep{vyas2024soap,gupta2018shampoo,ma2024swan,glentis2025scale}. However, these methods usually enforce a uniform global strategy, overlooking the potential benefits of block-wise configuration.

The feasibility of block-wise optimization is well supported. Conceptually, Adam's per-parameter scaling can be viewed as diagonal preconditioning and is naturally compatible with block-wise parameter grouping, motivating the idea that different blocks may benefit from different update rules and precisions. Empirically, layer-wise adaptation methods (LARS, LAMB) confirm that heterogeneous update scaling is stable \citep{You2018LARS,You2020LAMB}. While recent hybrid recipes (e.g., Adam-mini, Muon, and COSMOS) further validate the benefits of mixing update rules, they typically rely on static, manually designed heuristics \citep{zhang2024adammini,liu2025muon,cosmos2025}. BAOC addresses this limitation by formulating optimizer configuration as a systematic, data-driven resource allocation problem under given memory and time budgets.

\Needspace{4\baselineskip}

\section{Mechanism of BAOC}
\label{sec:method}

BAOC first splits the model parameters into $N$ blocks, and then follows four steps:
(1) sparse gradient sampling to update statistics;
(2) metric computation;
(3) solving a budget-constrained mixed-integer linear program (MILP);
and (4) applying the configurations. The process is given in Algorithm \ref{alg:baoc}.

\begin{algorithm}[ht]
\caption{BAOC: Budget-Aware Block-wise Optimizer Configuration}
\label{alg:baoc}
\begin{algorithmic}[1]
\STATE \textbf{Input:} Memory budget $B_{\mathrm{mem}}$; time budget $B_{\mathrm{time}}$; user weights $W$; baseline optimizer $\mathcal{O}_{\mathrm{base}}$; candidate configurations $\mathcal{C}$
\STATE \textbf{Output:} Block-wise optimizer configuration $\{c_i\}_{i=1}^N$ and configured optimizer $\mathcal{O}(\{c_i\})$

\STATE \textbf{Warmup and diagnostics:}
\STATE Run a warmup phase with $\mathcal{O}_{\mathrm{base}}$ and collect gradient streams using sparse coordinate sampling
\FOR{each block $i=1,\ldots,N$}
    \STATE Compute block-level metrics from sampled gradients
    \STATE Normalize metrics and form per-block risk signals using $W$
\ENDFOR

\STATE \textbf{Allocation:}
\STATE Construct the MILP in Eq.~\ref{eq:baoc_milp} using budgets $(B_{\mathrm{mem}}, B_{\mathrm{time}})$ and risks $\{R_i(\cdot)\}$
\STATE Solve the MILP to obtain selected configurations $\{c_i\}_{i=1}^N$
\STATE Instantiate the training optimizer $\mathcal{O}(\{c_i\})$ with the selected block-wise configurations

\STATE \textbf{Training:}
\FOR{$t=1, \ldots, T_{\mathrm{end}}$}
    \STATE Train one step with $\mathcal{O}(\{c_i\})$
\ENDFOR
\end{algorithmic}
\end{algorithm}

\subsection{Setup and Definition}
\label{sec:formulation}

We formalize the inputs, outputs, budgets and basic objects used by BAOC.
Let the trainable parameters be partitioned into $N$ disjoint blocks
$\mathcal{B}=\{\mathcal{B}_i\}_{i=1}^N$, where block $i$ contains $d_i$ parameters.

\paragraph{Inputs and Outputs.}
BAOC takes as input (i) lightweight block-wise gradient observations and
(ii) global system budgets $B_{\mathrm{mem}}$ and $B_{\mathrm{time}}$.

To reduce overhead, we do not process full gradients.
For each block $i$, we sample a subset of coordinates $\Omega_i \subseteq \{1,\ldots,d_i\}$
with $|\Omega_i|=\lceil s d_i\rceil$ (default $s=0.1\%$).%
\footnote{In our default implementation, $\Omega_i$ is fixed after sampling once.}
At step $t$, let $g_{i,t}\in\mathbb{R}^{d_i}$ be the block gradient, and record the sparse sample
$g_{i,t}^{\Omega}:=g_{i,t}[\Omega_i]$.
BAOC computes a block-wise metrics vector $z_i$ from the sampled streams $\{g_{i,t}^{\Omega}\}_t$ (e.g., via exponential moving averages (EMA)),
and outputs an assignment $i \mapsto c_i^\star$ that maps each block to a configuration in $\mathcal{C}$.

\paragraph{StateMem Definition.}
StateMem quantifies the \emph{persistent optimizer-owned state buffers} required to update parameters.
Let $\Theta$ be the set of trainable parameters. For a configuration $c$ and parameter $\theta \in \Theta$,
let $\mathcal{S}_c(\theta)$ denote the set of persistent state tensors maintained by the optimizer for $\theta$
(e.g., moments $m,v$, factored statistics, or other long-lived accumulators).
The total optimizer-state memory is
\[
\mathrm{StateMem}(c)=\sum_{\theta\in\Theta}\sum_{s\in\mathcal{S}_c(\theta)} |\theta_s|\cdot b(s),
\]
where $|\theta_s|$ is the element count of state tensor $s$, and $b(s)$ is the bytes per element determined by
the chosen precision (e.g., $4$ for FP32, $2$ for FP16/BF16, $1$ for INT8).
This metric \emph{strictly} excludes model parameters (including any auxiliary FP32 \emph{master weights}),
gradients, activations, temporary workspace/scratch buffers, and any additional parameter copies used for mixed precision
or evaluation (e.g., FP32 \emph{shadow} / EMA weights).
For the MILP, the block-wise memory cost is the restriction of the above sum to a block:
\[
M_{i,c}=\sum_{\theta\in\mathcal{B}_i}\sum_{s\in\mathcal{S}_c(\theta)} |\theta_s|\cdot b(s).
\]

\paragraph{Time Budget Definition.}
We use a relative update-time budget normalized to AdamW16.
For each block $i$, we define the baseline per-step update time of AdamW16 as unit time
($t_{i,\mathrm{AdamW16}} \equiv 1$).
For a candidate configuration $c$, the relative cost is
\[
r_{i,c} := \frac{t_{i,c}}{t_{i,\mathrm{AdamW16}}}.
\]
Here $t_{i,c}$ is obtained by a lightweight one-step timing measurement of updating block $i$ with configuration $c$
(on the target hardware and the same tensor shapes), and we reuse these ratios as an approximate cost model.
We approximate the per-step update-time cost of an allocation by the average of selected ratios:
\[
\tilde{T} := \frac{1}{N}\sum_i r_{i,c_i}.
\]
We use a conservative guardrail $\tilde{T} \le B_{\mathrm{time}}$ (default $B_{\mathrm{time}}=1.3$), i.e., the
aggregate update-time cost is estimated to be no more than $1.3\times$ the AdamW16 baseline under this proxy.
Note that $B_{\mathrm{time}}$ is an \emph{estimate-based} budget: due to runtime variability, it need not be satisfied
exactly at every step, but it helps avoid allocations with clearly excessive update overhead.
The timing overhead is negligible as it is performed only once during warmup.

\paragraph{Configuration Space.}
We define the candidate configuration space $\mathcal{C}$ using mechanism switches and state precision.
For applying configuration $c\in\mathcal{C}$ to block $i$, let $M_{i,c}$ be the required optimizer-state memory
and $t_{i,c}$ be the estimated per-step update-time cost.

\begin{definition}[Configuration]
\label{def:config}
A candidate configuration is a tuple
$c=(y_{\mathrm{a}},y_{\mathrm{m}},y_{\mathrm{d}},y_{\mathrm{f}},b)\in\{0,1\}^4\times\{32,16,8\}$,
where $y_{\mathrm{a}},y_{\mathrm{m}},y_{\mathrm{d}},y_{\mathrm{f}}$ indicate adaptive scaling, momentum,
decoupled weight decay, and state factorization, respectively, and $b$ is the state bit-width.
\end{definition}

\begin{remark}[Scope]
    In this work, we focus on the allocation mechanism under fixed budgets. While online replanning is possible, it requires optimizer-state migration and may introduce transient memory overhead; we leave a detailed study of replanning strategies to future work.
\end{remark}

\subsection{Optimization Model}

We formulate the allocation task as a MILP problem.

\paragraph{Variables and Objective.}
For each block $i$ and candidate configuration $c$, we introduce a binary variable
$x_{i,c}\in\{0,1\}$ indicating whether block $i$ selects configuration $c$.
BAOC minimizes a surrogate mismatch risk under global memory and update-time budgets.
To discourage overly aggressive compression when the budget allows, we define
\begin{equation}\label{eq:agg}
\mathrm{Agg}(c)
= (1-y_{\mathrm a})+(1-y_{\mathrm m})+(1-y_{\mathrm d})+y_{\mathrm f}
+{32}/{b}-1 .
\end{equation}
This term penalizes dropping adaptive scaling, momentum, or decoupled weight decay, using factorized states, and reducing state precision.
For compactness, define
\[
\Phi_{i,c}
:= R_i(c;z_i)+\gamma\,\mathrm{Agg}(c),
\]
where $\gamma=0.1$ by default.

\paragraph{The MILP Model.}
Given memory budget $B_{\mathrm{mem}}$ and update-time budget $B_{\mathrm{time}}$, BAOC solves
\begin{equation}
\label{eq:baoc_milp}
\begin{aligned}
\min_{x}\quad
& \sum_{i=1}^{N}\sum_{c\in\mathcal C} \Phi_{i,c}x_{i,c} \\
\mathrm{s.t.}\quad
& \sum_{c\in\mathcal C}x_{i,c}=1,\quad i=1,\ldots,N,\\
& \sum_{i=1}^{N}\sum_{c\in\mathcal C} M_{i,c}x_{i,c}\le B_{\mathrm{mem}},\\
& \frac{1}{N}\sum_{i=1}^{N}\sum_{c\in\mathcal C} r_{i,c}x_{i,c}\le B_{\mathrm{time}},\\
& x_{i,c}\in\{0,1\},\quad i=1,\ldots,N,\ c\in\mathcal C .
\end{aligned}
\end{equation}

The first constraint assigns exactly one configuration to each block.
The second and third constraints enforce the optimizer-state memory budget and the coarse update-time budget, respectively.
Here $r_{i,c}=t_{i,c}/t_{i,\mathrm{AdamW16}}$ is a relative update-time estimate, used only as a guardrail to avoid clearly slow allocations rather than as an exact predictor of end-to-end wall-clock time.
After solving Eq.~\eqref{eq:baoc_milp}, BAOC selects the unique configuration
$c_i^\star$ satisfying $x_{i,c_i^\star}^\star=1$ for each block.

\paragraph{Human Constraints.}
BAOC allows for the integration of domain priors through customizing constraints. We implement this via two mechanisms:

(1) \textit{Hard constraints}, which strictly exclude a risky candidate set $\mathcal{C}_{\mathrm{risk}}$, are enforced by setting $\sum_{c \in \mathcal{C}_{\mathrm{risk}}} x_{i,c} = 0$;

(2) \textit{Soft preferences}, which bias the selection toward a favored set $\mathcal{C}_{\mathrm{pref}}$, are injected by modifying the risk:
$$
R'_i(c;z_i)
= R_i(c;z_i)
-\lambda_{\mathrm{pref}}\mathbb{I}(c\in\mathcal{C}_{\mathrm{pref}}),
$$
where $\lambda_{\mathrm{pref}}$ serves as the weighting coefficient for the preference strength.

\begin{remark}[MILP solve time.]
Our allocation MILP is small-scale (tens of candidate configurations over tens of blocks) with simple linear budget constraints, and can be solved easily by modern MILP solvers (e.g., Gurobi) in less than 1 second. Therefore, the MILP solve time is negligible compared to training.
\end{remark}

\subsection{Gradient Metrics and Risk Construction}
\label{sec:baoc_diag}

BAOC relies on interpretable statistical metrics to map gradient behaviors to optimizer requirements. 
All metrics are computed efficiently using the sparse sampled gradients $g^{\Omega}_{i,t}$ defined in Sec.~\ref{sec:formulation}. 
Throughout this section, we use $\odot$ to denote the element-wise product and $\varepsilon$ as a small constant for numerical stability.
Unless stated otherwise, the sampled index set $\Omega_i$ is fixed for each block during warmup and training.

\paragraph{Moments.}
We maintain EMA for the first moment $m_{i,t}$ and second moment $v_{i,t}$ in block $i$ at step $t$:
\begin{equation}
\label{eq:moments}
\begin{aligned}
\hat m_{i,t} \leftarrow \beta_m \hat m_{i,t-1} + (1-\beta_m)\,g^{\Omega}_{i,t},\quad
\hat v_{i,t} \leftarrow \beta_v \hat v_{i,t-1} + (1-\beta_v)\,(g^{\Omega}_{i,t} \odot g^{\Omega}_{i,t}),
\end{aligned}
\end{equation}
where $\beta_m, \beta_v \in (0,1)$ are decay rates (defaulting to standard Adam values).

\paragraph{Anisotropy.} 
To determine the need for adaptive scaling, we measure the anisotropy of the second moment using quantile ratios:
\begin{equation}
\label{eq:anisotropy}
A_{i,t} = \log\frac{Q_{0.9}(\hat v_{i,t})+\varepsilon}{Q_{0.1}(\hat v_{i,t})+\varepsilon},
\end{equation}
where $Q_q(\cdot)$ denotes the $q$-th quantile function of the vector elements. High anisotropy suggests a wide range of gradient scales, implying a strong benefit from adaptive methods.

\paragraph{Dynamics.} 
To assess the need for momentum, we evaluate direction stability $\bar\rho_{i,t}$ and the signal-to-noise ratio (SNR) $\rho^{\mathrm{snr}}_{i,t}$:
\begin{equation}
\label{eq:rho_bar}
\begin{aligned}
    \bar\rho_{i,t} \leftarrow \beta_\rho \bar\rho_{i,t-1} + (1-\beta_\rho) \frac{\langle g^{\Omega}_{i,t}, g^{\Omega}_{i,t-1}\rangle}{\|g^{\Omega}_{i,t}\| \|g^{\Omega}_{i,t-1}\|+\varepsilon},\quad
\rho^{\mathrm{snr}}_{i,t} = \frac{\|\hat m_{i,t}\|^2}{\|\hat v_{i,t}\|_1+\varepsilon}.
\end{aligned}
\end{equation}
Here, $\langle \cdot, \cdot \rangle$ is the inner product, $\|\cdot\|$ is the $L_2$ norm, $\|\cdot\|_1$ is the $L_1$ norm, and $\beta_\rho$ is the smoothing factor for the stability metric.

\paragraph{Distortion.} 
Decoupled weight decay is critical when adaptive scaling distorts the gradient direction relative to the parameter weights. We measure this distortion $C_{i,t}$:
\begin{equation}
\label{eq:distortion}
C_{i,t} = \frac{\|(p_{i,t}/\bar p_{i,t}-\mathbf{1})\odot \theta^{\Omega}_{i,t}\|}{\|\theta^{\Omega}_{i,t}\|+\varepsilon},
\end{equation}
where $p_{i,t} = (\sqrt{\hat v_{i,t}}+\varepsilon)^{-1}$ is the preconditioner, $\bar p_{i,t}$ is its mean value, $\mathbf{1}$ is a vector of ones, and $\theta^{\Omega}_{i,t}$ represents the sampled model parameter values.

\paragraph{Structure.} 
For matrix blocks, we test if the second moment approximates a rank-1 outer product. Let $G^{\Omega}_{i,t}$ be the matrix form of the sampled gradients. We update a structural second moment $S_{i,t}$ and compute the relative approximation error $F_{i,t}$:
\begin{equation}
\label{eq:structure_res}
S_{i,t} \leftarrow \text{EMA}(G^{\Omega}_{i,t} \odot G^{\Omega}_{i,t}), \quad
F_{i,t} = \frac{\|S_{i,t} - \tilde S_{i,t}\|_F}{\|S_{i,t}\|_F+\varepsilon},
\end{equation}
where $\tilde S_{i,t}$ is the best rank-1 approximation derived from row and column means of $S_{i,t}$, and $\|\cdot\|_F$ denotes the Frobenius norm.

\paragraph{Precision Sensitivity.} 
We measure robustness to quantization by simulating low-precision updates. 
Let $\mathcal{Q}_b(\cdot)$ be a standard quantizer for bit-width $b$. We compute the cosine similarity $Q_{i,t}(b)$ between a full-precision update direction $u^{32}$ and a quantized update direction $u^{b}$ (i.e., the parameter update direction applied to $\theta$):
\begin{equation}
\label{eq:precision_cos}
Q_{i,t}(b) = \frac{\langle u^{32}_{i,t}, u^{b}_{i,t}\rangle}{\|u^{32}_{i,t}\| \|u^{b}_{i,t}\|+\varepsilon}.
\end{equation}

\paragraph{Risk Construction.}
We normalize the raw metrics above into standard scores: $s_{A,i}, s_{M,i}, \tilde C_i, s_{F,i},$ and $\ell_{Q,i}(b) = -\log(Q_{i,t}(b))$ (the details of the normalization are provided in Appendix ~\ref{app:norm}).
We summarize each metric by its final value at the end of warmup, and treat the resulting block-wise diagnostics as $z_i$ in the allocation stage.
The final risk function $R_i(c; z_i)$ is a linear combination:
\begin{equation}
\label{eq:default_risk_linear}
R_i(c;z_i)
= s_{A,i}(1-y_{\mathrm a})
+ s_{M,i}(1-y_{\mathrm m})
+ \tilde C_i(1-y_{\mathrm d})
+ s_{F,i}y_{\mathrm f}
+ \ell_{Q,i}(b).
\end{equation}
BAOC supports user-defined weights during risk quantification, but we do not apply differential weighting in the default configuration.

\begin{remark}[Decoupled mechanism-wise risk proxy]
The linear risk term is a pragmatic surrogate for budgeted allocation. 
Its design is motivated by a common pattern in modern large-scale optimization: effective optimizers often improve training by separating different mechanisms, including adaptive scaling in Adam~\citep{Kingma2015Adam}, decoupled weight decay in AdamW~\citep{Loshchilov2019AdamW}, curvature- or Fisher-motivated preconditioning~\citep{gupta2018shampoo}, factorized or low-rank states~\citep{shazeer2018adafactor,pmlr-v235-zhao24s}, quantized optimizer states~\citep{DBLP:conf/iclr/DettmersLSZ22,zhang2024qgalore}, and Muon-style orthogonalized updates~\citep{jordan2024muon,liu2025muon}. 
We therefore use anisotropy, direction stability, distortion, structure residual, and precision sensitivity as block-level proxies for the need to preserve different optimizer mechanisms under a memory budget. 
The linear form ignores interaction effects among mechanisms and across blocks; modeling such couplings is left to future work.
\end{remark}

\section{Experiments}
\label{sec:exp}

We evaluate BAOC on vision, language, and diffusion training, with the goal of assessing whether block-wise optimizer configuration can retain training quality under externally specified optimizer-state memory budgets.
Across all settings, BAOC operates at the parameter-block level and assigns each block an optimizer configuration from a candidate set under global memory and time constraints.

We organize the experiments into three parts:
(1) a fixed-budget allocation study under the default StateMem budget;
(2) a memory-budget sweep under different externally specified StateMem constraints;
and (3) a longer-schedule language-modeling comparison with recent hybrid optimizers.
Additional diagnostics, including sampling ratio, regularization strength, anchor perturbation, block partitioning, human constraints, and online reallocation, are reported in Appendix~\ref{app:add-exp}.

\subsection{Experimental setup}
\label{sec:exp-setup}

\textbf{Benchmarks.}
We consider a diverse set of workloads:
(1) ViT for image classification on ImageNet-1K~\citep{dosovitskiy2021vit,deng2009imagenet};
(2) GPT2-small for text generation on Alpaca and GSM8K~\citep{radford2019gpt2,taori2023alpaca,cobbe2021gsm8k};
(3) T5-base for text generation on Alpaca and GSM8K~\citep{raffel2020t5,taori2023alpaca,cobbe2021gsm8k};
(4) a diffusion UNet model~\citep{ronneberger2015unet,ho2020ddpm};
and (5) Llama-3.2-1B and Llama-3.2-3B for text generation on Alpaca and GSM8K, following the Llama-3 family setup~\citep{touvron2023llama,llamateam2024llama3}.

Unless stated otherwise, all methods are evaluated under the same task setup, including dataset split, number of training epochs/steps, batch size, and model architecture.
For each optimizer baseline, we use the learning-rate schedule and optimizer-specific hyperparameters recommended in the original paper or official implementation.
The main controlled experiments are conducted on a cluster of three NVIDIA RTX PRO 6000 Blackwell GPUs, each with 96GB memory.

\textbf{Baselines.}
We compare BAOC with representative non-BAOC baselines, including AdamW8 with 8-bit optimizer states~\citep{DBLP:conf/iclr/DettmersLSZ22}, AdamW16~\citep{Loshchilov2019AdamW}, Adam-mini~\citep{zhang2024adammini,han2025qadammini}, GaLore-AdamW8, GaLore-AdamW32, GaLore-Adafactor~\citep{pmlr-v235-zhao24s,shazeer2018adafactor}, and advanced SPAM~\citep{huang2025spam}.
These baselines cover full-state adaptivity, quantized states, memory-efficient designs, spike-aware stabilization, subspace-update training, and factorized second-moment statistics.
For the longer-schedule comparison in Sec.~\ref{sec:exp-longer-schedule}, we additionally include recent hybrid optimizers, including Muon~\citep{liu2025muon} and COSMOS~\citep{cosmos2025}.

\textbf{BAOC configuration.}
We use AdamW16 as the reference baseline and define the optimizer-state memory budget as
$B_{\mathrm{mem}}=\rho\cdot\mathrm{StateMem}(\text{AdamW16})$,
where $\rho$ is the budget ratio.
Unless stated otherwise, the default BAOC setting uses $\rho=0.5$.
We use a structure-guided and statistic-refined partitioning strategy: model structural boundaries define candidate units, while warmup diagnostics and a minimum-size constraint determine the final blocks.
Details are provided in Appendix~\ref{app:partition}.
BAOC samples gradients within each block to estimate the diagnostics in Sec.~\ref{sec:baoc_diag}, constructs block--configuration risks, and solves the MILP in Eq.~\eqref{eq:baoc_milp}.
The configuration set includes AdamW, Adam, SGD, SGD with momentum, SGDW, SGDW with momentum, and Adafactor, each implemented with 8/16/32-bit optimizer states.
We set $\gamma=0.1$ in Eq.~\eqref{eq:agg} by default.

\textbf{Metrics.}
We report the task metric associated with each workload: test accuracy for ViT, training loss for diffusion UNet, and test perplexity (PPL) for language modeling tasks.
We also report StateMem, the persistent optimizer-state memory directly controlled by BAOC, and wall-clock time.
When relevant, we additionally report PeakMem, the peak end-to-end GPU memory footprint observed during training.
StateMem evaluates whether the allocation satisfies the optimizer-state budget, while PeakMem reflects the full training memory pressure, including activations, gradients, and temporary buffers.
For Experiments 1 and 2, we run at least three random seeds and report the mean and standard deviation.
For the longer-schedule comparison, we report the final values from the corresponding long-run setting due to its substantially higher computational cost.
Complete numerical results are provided in the appendix.

\subsection{Experiment 1: Fixed-budget allocation}
\label{sec:exp-fixed-budget}

We first evaluate BAOC under a fixed optimizer-state memory budget.
For each workload, BAOC is constrained to use at most $50\%$ of the optimizer-state memory of AdamW16, i.e., $\rho=0.5$.
After a short warmup for gradient diagnostics, BAOC solves the MILP in Eq.~\eqref{eq:baoc_milp} and trains with the selected block-wise configurations.

\begin{figure}[ht]
  \centering
  \includegraphics[width=\textwidth]{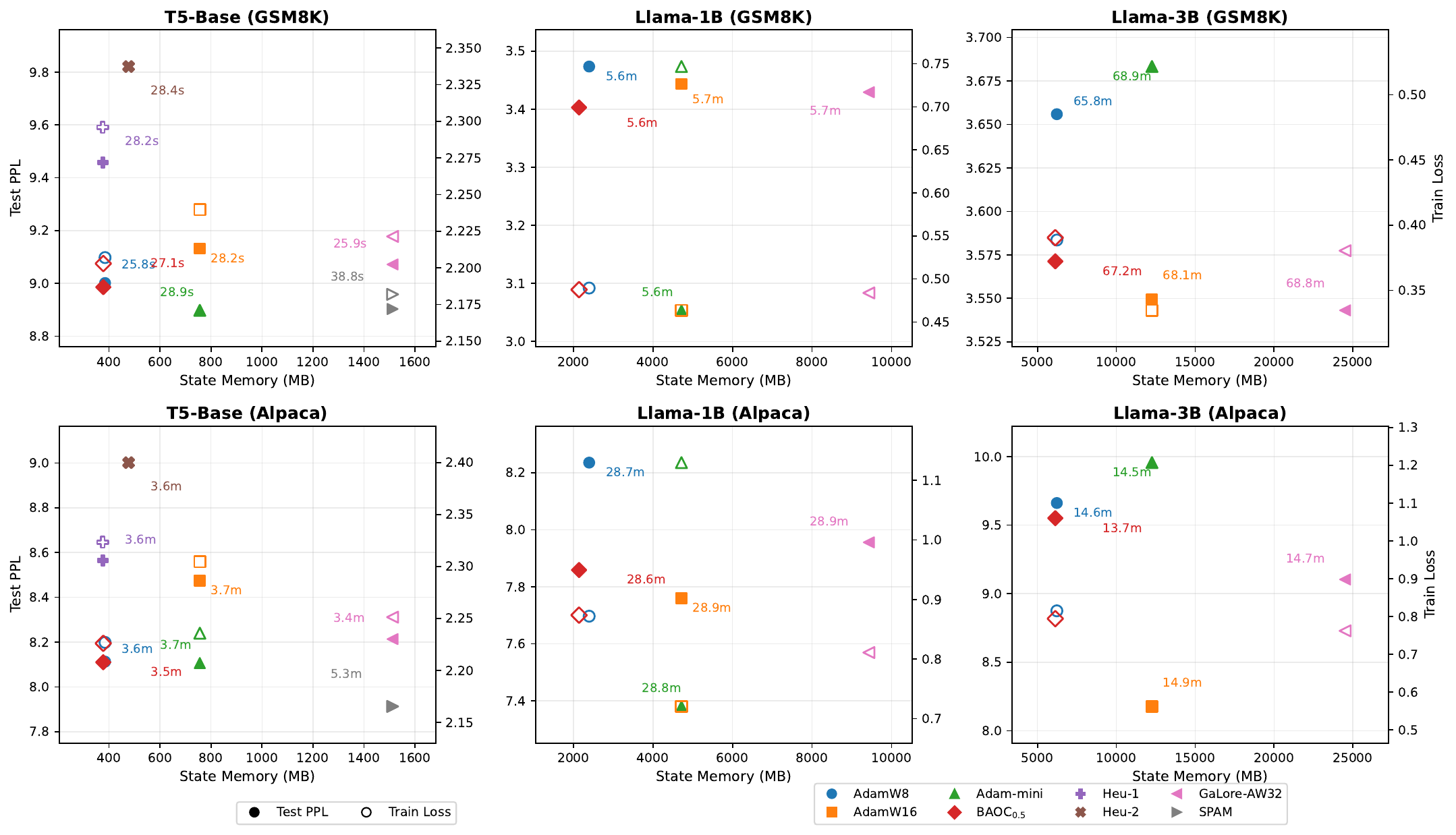}
  \caption{
  Fixed-budget allocation under the default StateMem budget.
  Each subplot corresponds to one language workload.
  Filled markers report test perplexity on the left y-axis, and hollow markers report training loss on the right y-axis.
  Runtime is annotated near each method.
  BAOC$_{0.5}$ denotes BAOC with $\rho=0.5$.
  }
  \label{fig:exp1-tradeoff}
\end{figure}

Figure~\ref{fig:exp1-tradeoff} summarizes the fixed-budget results.
Across T5-base and Llama workloads, BAOC$_{0.5}$ satisfies the StateMem constraint and provides competitive test PPL while using substantially less optimizer-state memory than full-state baselines.
The annotated runtimes show that the memory reduction does not require a systematic increase in wall-clock training time.
Complete numerical results, including standard deviations, are provided in Appendix~\ref{app:fixed-budget-full}.

\subsection{Experiment 2: Memory-budget sweep}
\label{sec:exp-budget-sweep}

We next examine BAOC under different externally specified optimizer-state memory budgets.
We vary the budget ratio $\rho$ relative to the optimizer-state memory of AdamW16 and run BAOC with
$\rho\in\{0.3,0.4,0.5,0.55,0.6,0.7,0.8,1.0,1.2\}$ whenever applicable.
For each budget, BAOC performs allocation under the corresponding StateMem constraint and trains with the resulting block-wise configurations.

\begin{figure}[ht]
  \centering
  \includegraphics[width=\textwidth]{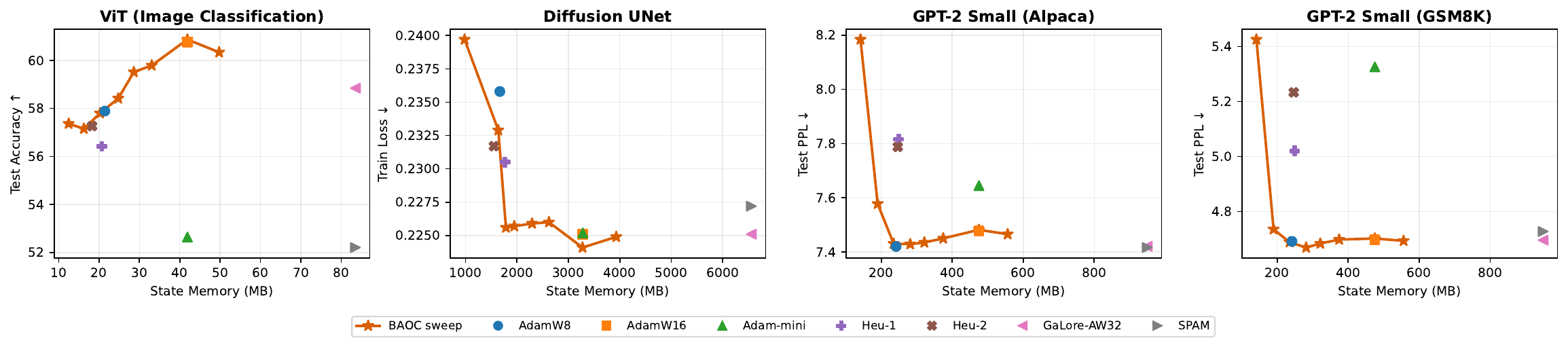}
  \caption{
  Memory-budget sweep under different StateMem constraints.
  The orange star curve shows BAOC allocations under different budget ratios $\rho$, while isolated markers denote representative baseline optimizers.
  The y-axis reports the task metric for each workload:
  test accuracy for ViT, training loss for diffusion UNet, and test perplexity for GPT-2.
  Higher is better for accuracy, while lower is better for loss and perplexity.
  }
  \label{fig:exp2_tradeoff}
\end{figure}

Figure~\ref{fig:exp2_tradeoff} reports the empirical outcomes across image classification, diffusion training, and autoregressive language modeling.
As the budget increases, BAOC can select less aggressive configurations, but the final task metric is not necessarily monotone because the allocation objective minimizes a surrogate block-level mismatch risk rather than the validation metric directly.
Across workloads, BAOC provides useful budget-conditioned allocations over a range of memory constraints and remains competitive with homogeneous global baselines under constrained StateMem budgets.
Complete numerical results are reported in Appendix~\ref{app:budget-sweep-full}.

\subsection{Experiment 3: Longer-schedule comparison with hybrid optimizers}
\label{sec:exp-longer-schedule}

We further evaluate BAOC on longer language-modeling runs and compare it with recent hybrid optimizers.
We use GPT2-small and T5-base on Alpaca and GSM8K, with 200 epochs for GPT2-small and 100 epochs for T5-base in a 3-GPU setup.
We compare Muon, COSMOS, Adam-mini, AdamW16, and BAOC$_{0.5}$, and report final test PPL, training time, and StateMem.

\begin{small}
\begin{table}[ht]
\centering
\small
\setlength{\tabcolsep}{4pt}
\caption{
Longer-schedule comparison with recent hybrid optimizers.
Lower test PPL, training time, and StateMem are better.
Time is reported in minutes and StateMem in MB.
}
\label{tab:longer-schedule}
\begin{tabular}{ccccc}
\toprule
\textbf{Task} & \textbf{Method} & \textbf{Test PPL} $\downarrow$ & \textbf{Time (min)} $\downarrow$ & \textbf{StateMem (MB)} $\downarrow$ \\
\midrule
\multirow{5}{*}[0pt]{\centering GPT2-small/Alpaca}
& Muon       & 8.594 & 251.0 & 625 \\
& COSMOS     & 8.059 & 415.0 & 660 \\
& Adam-mini  & 7.600 & 305.0 & 475 \\
& AdamW16    & 7.442 & 348.0 & 475 \\
& BAOC$_{0.5}$ & \textbf{7.391} & \textbf{182.0} & \textbf{235} \\
\midrule
\multirow{5}{*}[0pt]{\centering GPT2-small/GSM8K}
& Muon       & 6.238 & 35.8 & 625 \\
& COSMOS     & 4.737 & 40.3 & 660 \\
& Adam-mini  & 4.995 & \textbf{30.2} & 475 \\
& AdamW16    & 4.615 & 31.5 & 475 \\
& BAOC$_{0.5}$ & \textbf{4.598} & 30.8 & \textbf{235} \\
\midrule
\multirow{5}{*}[0pt]{\centering T5-base/Alpaca}
& Muon       & 10.619 & 313.0 & 945 \\
& COSMOS     & 8.504  & 481.0 & 858 \\
& Adam-mini  & 8.107  & 224.0 & 756 \\
& AdamW16    & 8.475  & 224.0 & 756 \\
& BAOC$_{0.5}$ & \textbf{8.100} & \textbf{211.0} & \textbf{378} \\
\midrule
\multirow{5}{*}[0pt]{\centering T5-base/GSM8K}
& Muon       & 13.049 & 39.2 & 945 \\
& COSMOS     & 9.921  & 61.3 & 858 \\
& Adam-mini  & \textbf{8.898} & 28.9 & 756 \\
& AdamW16    & 9.312  & 28.2 & 756 \\
& BAOC$_{0.5}$ & 9.137 & \textbf{22.8} & \textbf{357} \\
\bottomrule
\end{tabular}
\end{table}
\end{small}

Table~\ref{tab:longer-schedule} shows that BAOC$_{0.5}$ remains competitive under longer schedules.
It obtains the best PPL on both GPT2-small tasks and on T5-base/Alpaca, while using roughly half the StateMem of AdamW16.
On T5-base/GSM8K, Adam-mini attains a slightly lower PPL, but BAOC uses less StateMem and the lowest training time.
These results support BAOC as a fixed-budget allocation mechanism rather than a method that must dominate every baseline in final PPL.

\subsection{Additional diagnostics}
Appendix~\ref{app:add-exp} reports additional diagnostics on sampling ratio, Agg coefficient $\gamma$, normalization-anchor perturbation, human preference constraints, block partitioning, and online reallocation.
Only settings that change the selected allocation relative to BAOC$_{0.5}$ are listed; unchanged settings are omitted as redundant.

\section{Discussion}\label{sec:discussion}

BAOC supports the hypothesis that optimizer configuration can be treated as an allocatable resource, and provides an initial empirical link between optimizer update mechanisms and block-wise gradient dynamics.
The current framework is designed as a fixed-budget conditional allocator: given externally specified memory and time budgets, BAOC selects block-wise optimizer configurations by minimizing a surrogate mismatch risk.
It does not aim to automatically choose the globally best budget or construct a strict Pareto frontier over memory and final validation performance.

There are several limitations and possible extensions.
First, the current risk model is additive across blocks and mechanisms, and does not directly optimize the final validation metric.
Modeling cross-block interactions or learning a more predictive allocation risk remains an important direction.
Second, the current configuration space covers common optimizer mechanisms and state precisions, but can be further expanded to include emerging optimizers and broader quantizable states.
Third, online re-planning requires smoother optimizer-state transitions; as shown in our diagnostics, reallocation without proper state inheritance can degrade training quality.
Fourth, BAOC currently provides simple interfaces for human constraints, but more fine-grained domain priors may further improve allocation reliability.
Finally, integrating BAOC with distributed training frameworks such as ZeRO and FSDP is necessary for evaluating its behavior in large-scale sharded training.

Overall, BAOC represents a block-wise optimizer allocation philosophy.
The MILP formulation used in this work is a simple and practical instantiation, while the broader framework remains extensible to richer risk models, larger configuration spaces, and more system-aware allocation constraints.

\section*{Impact Statement}
BAOC aims to reduce optimizer-state memory and improve the accessibility and efficiency of model training. Its potential positive impacts include lowering hardware requirements, reducing training cost, and improving resource utilization. At the same time, more efficient training methods may also lower the barrier to training larger models, including models that could be misused. This work releases optimizer-allocation code rather than new datasets or pretrained generative models. Users should follow the accompanying license and the terms of use of all datasets, models, and dependencies.

\bibliography{example_paper}
\bibliographystyle{plainnat}

\newpage

\appendix
\section{Normalization and risk signals}
\label{app:norm}

This appendix specifies the default normalization functions used in our experiments.
All functions are simple, deterministic, and require no tuning from validation data.
They map raw diagnostics into comparable need or risk signals that are then used in
Eq.~\eqref{eq:default_risk_linear}.

\subsection{Common preprocessing}
\label{app:norm_common}

We apply the following preprocessing to all blocks.

\begin{itemize}
\item Numerical constant: $\varepsilon=10^{-12}$.
\item Clipping for cosine based scores:
\begin{equation}
\label{eq:clip_Q}
Q_i(b)\leftarrow \mathrm{clip}\bigl(Q_i(b),\,\varepsilon,\,1\bigr),
\qquad b\in\{32,16,8\}.
\end{equation}
\item If $\rho_{i,t}$ is NaN due to zero norms, set $\rho_{i,t}=0$.
\item Raw diagnostics are computed from EMA as described in
Sec.~\ref{sec:baoc_diag}.
\end{itemize}

\subsection{Geometry signal from anisotropy}
\label{app:norm_geom}

We use the raw anisotropy score $A_i$ from Eq.~\eqref{eq:anisotropy}.
Because $A_i=\log(Q_{0.9}/Q_{0.1})$ has an absolute scale, we avoid dataset level
rank based normalization.

We choose two fixed anchors:
\begin{equation}
\label{eq:A_anchors}
A_{\mathrm{low}}=\log 2,
\qquad
A_{\mathrm{high}}=\log 10.
\end{equation}
They correspond to a $2\times$ and $10\times$ quantile ratio.

The geometry need signal is
\begin{equation}
\label{eq:sA}
s_{A,i}=
\mathrm{clip}\!\left(
\frac{A_i-A_{\mathrm{low}}}{A_{\mathrm{high}}-A_{\mathrm{low}}},
\,0,\,1
\right).
\end{equation}

\subsection{Momentum need from direction and SNR}
\label{app:norm_mom}

Momentum is beneficial when gradient directions are stable and the accumulated
signal is strong relative to the second moment scale.
We combine direction stability $\bar\rho_i$ and
the magnitude proxy $\rho^{\mathrm{snr}}_i$ from Eq.~\eqref{eq:rho_bar}.

We use two anchors for direction stability:
\begin{equation}
\label{eq:rho_anchors}
\rho_{\mathrm{low}}=0.2,
\qquad
\rho_{\mathrm{high}}=0.6.
\end{equation}
The direction gate is
\begin{equation}
\label{eq:s_rho}
s_{\rho,i}=
\mathrm{clip}\!\left(
\frac{\bar\rho_i-\rho_{\mathrm{low}}}{\rho_{\mathrm{high}}-\rho_{\mathrm{low}}},
\,0,\,1
\right).
\end{equation}

We compress the SNR proxy by $\log(1+x)$ and map it to $[0,1]$ using two anchors:
\begin{equation}
\label{eq:snr_anchors}
\eta_{\mathrm{low}}=0,
\qquad
\eta_{\mathrm{high}}=2.
\end{equation}
The SNR gate is
\begin{equation}
\label{eq:s_snr}
s_{\mathrm{snr},i}=
\mathrm{clip}\!\left(
\frac{\log(1+\rho^{\mathrm{snr}}_i)-\eta_{\mathrm{low}}}{\eta_{\mathrm{high}}-\eta_{\mathrm{low}}},
\,0,\,1
\right).
\end{equation}

The default momentum need is the product
\begin{equation}
\label{eq:sM}
s_{M,i}=s_{\rho,i}\cdot s_{\mathrm{snr},i}.
\end{equation}
If the SNR proxy is not available or is unreliable in warmup, we use
$s_{M,i}=s_{\rho,i}$.

\subsection{Regularization distortion signal}
\label{app:norm_reg}

We use $C_i$ from Eq.~\eqref{eq:distortion}.
Its scale can be large, so we apply log compression:
\begin{equation}
\label{eq:Ctilde}
\tilde C_i=\log(1+C_i).
\end{equation}
We keep $\tilde C_i$ unbounded and let the weight $w_C$ absorb scale differences.

\subsection{Structure signal}
\label{app:norm_struct}

We use the structure residual $F_i$ from Eq.~\eqref{eq:structure_res}.
It is naturally in $[0,1]$ in most cases.
We clip it for safety:
\begin{equation}
\label{eq:sF}
s_{F,i}=\mathrm{clip}(F_i,0,1).
\end{equation}

\subsection{Precision risk from cosine similarity}
\label{app:norm_prec}

We use the cosine similarity $Q_i(b)$ from Eq.~\eqref{eq:precision_cos}.
We convert it into a risk that increases as alignment decreases:
\begin{equation}
\label{eq:lQ}
\ell_{Q,i}(b)=-\log\bigl(Q_i(b)+\varepsilon\bigr),
\qquad b\in\{32,16,8\}.
\end{equation}
This mapping amplifies small differences when $Q$ is close to $1$ and gives
$\ell_Q\to 0$ as $Q\to 1$.

\subsection{Default linear risk}
\label{app:norm_risk}

The final risk used in Eq.~\eqref{eq:default_risk_linear} is
\begin{equation} 
\label{eq:app_default_risk} 
\begin{aligned} R_i(c;z_i)= s_{A,i}\bigl(1-y_{\mathrm{a}}(c)\bigr) +s_{M,i}\bigl(1-y_{\mathrm{m}}(c)\bigr)+\tilde C_i\bigl(1-y_{\mathrm{d}}(c)\bigr) +s_{F,i} y_{\mathrm{f}}(c) +\ell_{Q,i}\bigl(b(c)\bigr). \end{aligned} 
\end{equation}


\section{Full Results for Fixed-Budget Allocation}
\label{app:fixed-budget-full}

\begin{center}
\scriptsize
\setlength{\tabcolsep}{2.0pt}
\setlength{\LTpre}{0pt}
\setlength{\LTpost}{0pt}
\renewcommand{\arraystretch}{0.98}

\begin{longtable}{@{}cccccc@{}}
\caption{Detailed fixed-budget results (mean $\pm$ std). State denotes optimizer-state memory in MB, and Time is in seconds. We report training loss and train/test PPL where applicable.}
\label{tab:fixed_budget_full}\\
\toprule
\textbf{Method} & \textbf{State} & \textbf{Loss} & \textbf{Time} & \textbf{Tr. PPL} & \textbf{Te. PPL} \\
\midrule
\endfirsthead

\toprule
\textbf{Method} & \textbf{State} & \textbf{Loss} & \textbf{Time} & \textbf{Tr. PPL} & \textbf{Te. PPL} \\
\midrule
\endhead

\midrule
\multicolumn{6}{c}{\textit{Continued on next page.}} \\
\endfoot

\bottomrule
\endlastfoot

\multicolumn{6}{c}{\textbf{Llama\_3.2\_1B\_alpaca}} \\
\midrule
Adam-mini      & $4715.5338 \pm 0.0000$ & $1.1298 \pm 0.0015$ & $1729.2948 \pm 30.9504$ & $2.2214 \pm 0.0056$ & $7.3804 \pm 0.4692$ \\
AdamW16        & $4714.5000 \pm 0.0000$ & $0.7207 \pm 0.0006$ & $1731.4853 \pm 22.3383$ & $1.5548 \pm 0.0018$ & $7.7593 \pm 0.0236$ \\
AdamW8         & $2394.5488 \pm 0.0000$ & $0.8721 \pm 0.0177$ & $1720.7466 \pm 28.2941$ & $1.7859 \pm 0.0040$ & $8.2353 \pm 0.0866$ \\
BAOC$_{0.5}$   & $2140.1338 \pm 0.0000$ & $0.8739 \pm 0.0139$ & $1717.8212 \pm 17.0548$ & $1.7823 \pm 0.0172$ & $7.8588 \pm 0.1373$ \\
GaLore-AF      & $3.4351 \pm 0.0000$    & $1.3680 \pm 0.0003$ & $1768.7772 \pm 17.2582$ & $3.7171 \pm 0.0006$ & $8.3177 \pm 0.0014$ \\
GaLore-AW32    & $9428.5000 \pm 0.0000$ & $0.8111 \pm 0.0011$ & $1733.7957 \pm 19.6231$ & $1.6861 \pm 0.0133$ & $7.9558 \pm 0.0494$ \\
\addlinespace

\multicolumn{6}{c}{\textbf{Llama\_3.2\_1B\_gsm8k}} \\
\midrule
Adam-mini      & $4715.5338 \pm 0.0000$ & $0.7468 \pm 0.0474$ & $338.5162 \pm 0.7952$ & $1.6950 \pm 0.0667$ & $3.0536 \pm 0.0296$ \\
AdamW16        & $4714.5000 \pm 0.0000$ & $0.4636 \pm 0.0038$ & $339.7833 \pm 1.3011$ & $1.3168 \pm 0.0094$ & $3.4436 \pm 0.0141$ \\
BAOC$_{0.5}$   & $2140.1338 \pm 0.0000$ & $0.4876 \pm 0.0039$ & $336.2259 \pm 0.2760$ & $1.3444 \pm 0.0061$ & $3.4030 \pm 0.0398$ \\
GaLore-AF      & $3.4351 \pm 0.0000$    & $0.9615 \pm 0.0006$ & $344.5018 \pm 0.4129$ & $2.5308 \pm 0.0014$ & $2.7346 \pm 0.0016$ \\
GaLore-AW32    & $9428.5000 \pm 0.0000$ & $0.4838 \pm 0.0022$ & $339.3417 \pm 1.7266$ & $1.3357 \pm 0.0034$ & $3.4293 \pm 0.0113$ \\
AdamW8         & $2394.5488 \pm 0.0000$ & $0.4895 \pm 0.0006$ & $335.7520 \pm 1.3582$ & $1.3419 \pm 0.0036$ & $3.4735 \pm 0.0384$ \\
\addlinespace

\multicolumn{6}{c}{\textbf{Llama\_3.2\_3B\_alpaca}} \\
\midrule
Adam-mini      & $12258.1823 \pm 0.0000$ & $1.2070 \pm 0.1133$ & $869.8237 \pm 4.3450$ & $2.2555 \pm 0.1746$ & $9.9557 \pm 0.4247$ \\
AdamW16        & $12256.3125 \pm 0.0000$ & $0.5622 \pm 0.0037$ & $895.5152 \pm 4.4552$ & $1.4073 \pm 0.0052$ & $8.1776 \pm 0.1670$ \\
AdamW8         & $6224.9395 \pm 0.0000$  & $0.8150 \pm 0.0113$ & $876.4541 \pm 4.4265$ & $1.6423 \pm 0.0002$ & $9.6611 \pm 0.1421$ \\
BAOC$_{0.5}$   & $6127.4092 \pm 0.0000$  & $0.7944 \pm 0.0012$ & $824.0374 \pm 4.1201$ & $1.6526 \pm 0.0007$ & $9.5502 \pm 0.1485$ \\
GaLore-AF      & $6.9551 \pm 0.0000$     & $1.1854 \pm 0.0004$ & $890.5379 \pm 4.2583$ & $2.9719 \pm 0.0004$ & $8.9715 \pm 0.1742$ \\
GaLore-AW32    & $24511.3125 \pm 0.0000$ & $0.7621 \pm 0.0109$ & $881.6705 \pm 4.4014$ & $1.6311 \pm 0.0092$ & $9.1036 \pm 0.1773$ \\
\addlinespace

\multicolumn{6}{c}{\textbf{Llama\_3.2\_3B\_gsm8k}} \\
\midrule
Adam-mini      & $12258.1823 \pm 0.0000$ & $0.5216 \pm 0.0319$ & $4135.7124 \pm 140.3410$ & $1.3837 \pm 0.0922$ & $3.6833 \pm 0.1574$ \\
AdamW16        & $12256.3125 \pm 0.0000$ & $0.3347 \pm 0.0027$ & $4085.5169 \pm 168.2340$ & $1.2031 \pm 0.0024$ & $3.5495 \pm 0.0591$ \\
AdamW8         & $6224.9395 \pm 0.0000$  & $0.3886 \pm 0.0084$ & $3945.8413 \pm 160.4061$ & $1.2522 \pm 0.0209$ & $3.6559 \pm 0.0609$ \\
BAOC$_{0.5}$   & $6127.4092 \pm 0.0000$  & $0.3904 \pm 0.0014$ & $4030.8570 \pm 157.7422$ & $1.2501 \pm 0.0156$ & $3.5713 \pm 0.0585$ \\
GaLore-AF      & $6.9551 \pm 0.0000$     & $0.8065 \pm 0.0002$ & $4191.0108 \pm 162.9693$ & $2.1099 \pm 0.0002$ & $2.4887 \pm 0.0622$ \\
GaLore-AW32    & $24511.3125 \pm 0.0000$ & $0.3804 \pm 0.0121$ & $4125.9537 \pm 165.6456$ & $1.2343 \pm 0.0015$ & $3.5431 \pm 0.0590$ \\
\addlinespace

\multicolumn{6}{c}{\textbf{t5\_base\_alpaca}} \\
\midrule
Adam-mini      & $756.2528 \pm 0.0000$  & $2.2358 \pm 0.0053$ & $223.7034 \pm 10.0226$ & $7.1087 \pm 0.0403$ & $8.1067 \pm 0.0252$ \\
AdamW16        & $756.3574 \pm 0.0000$  & $2.3045 \pm 0.0057$ & $224.3498 \pm 4.0162$  & $7.7071 \pm 0.0385$ & $8.4751 \pm 0.0290$ \\
AdamW8         & $384.6387 \pm 0.0000$  & $2.2269 \pm 0.0011$ & $217.4114 \pm 6.4462$  & $7.0098 \pm 0.0178$ & $8.1129 \pm 0.0041$ \\
BAOC$_{0.5}$   & $377.7700 \pm 0.0000$  & $2.2260 \pm 0.0032$ & $210.7029 \pm 7.0970$  & $7.0043 \pm 0.0216$ & $8.1101 \pm 0.0132$ \\
Heu-1          & $375.7593 \pm 0.0000$  & $2.3233 \pm 0.0015$ & $214.0850 \pm 11.8173$ & $7.7922 \pm 0.0080$ & $8.5644 \pm 0.0105$ \\
Heu-2          & $476.9957 \pm 0.0000$  & $2.3999 \pm 0.0030$ & $218.3272 \pm 10.3129$ & $8.4623 \pm 0.0372$ & $9.0016 \pm 0.0241$ \\
GaLore-AF      & $1.7240 \pm 0.0000$    & $2.4392 \pm 0.0013$ & $287.7375 \pm 9.2421$  & $8.9209 \pm 0.0082$ & $9.2030 \pm 0.0012$ \\
GaLore-AW32    & $1512.3574 \pm 0.0000$ & $2.2512 \pm 0.0028$ & $202.6089 \pm 5.5004$  & $7.2148 \pm 0.0277$ & $8.2136 \pm 0.0178$ \\
SPAM           & $1512.3574 \pm 0.0000$ & $2.1653 \pm 0.0016$ & $315.3652 \pm 23.2793$ & $6.4806 \pm 0.0112$ & $7.9128 \pm 0.0055$ \\
\addlinespace

\multicolumn{6}{c}{\textbf{t5\_base\_gsm8k}} \\
\midrule
Adam-mini      & $756.2528 \pm 0.0000$  & $2.1710 \pm 0.0185$ & $28.8764 \pm 0.5487$ & $7.5267 \pm 0.1652$ & $8.8979 \pm 0.0294$ \\
AdamW16        & $756.3574 \pm 0.0000$  & $2.2397 \pm 0.0032$ & $28.2331 \pm 1.3939$ & $8.2342 \pm 0.0211$ & $9.1318 \pm 0.0251$ \\
AdamW8         & $384.6387 \pm 0.0000$  & $2.2070 \pm 0.0032$ & $25.7640 \pm 0.3625$ & $7.8765 \pm 0.0401$ & $9.0013 \pm 0.0329$ \\
BAOC$_{0.5}$   & $377.7700 \pm 0.0000$  & $2.2029 \pm 0.0098$ & $27.0551 \pm 0.2333$ & $7.8627 \pm 0.1020$ & $8.9859 \pm 0.0457$ \\
Heu-1          & $375.7593 \pm 0.0000$  & $2.2958 \pm 0.0065$ & $28.2087 \pm 0.4501$ & $8.7887 \pm 0.1116$ & $9.4577 \pm 0.0931$ \\
Heu-2          & $476.9957 \pm 0.0000$  & $2.3373 \pm 0.0067$ & $28.4228 \pm 0.7139$ & $9.3485 \pm 0.0305$ & $9.8211 \pm 0.0510$ \\
GaLore-AF      & $1.7240 \pm 0.0000$    & $2.3827 \pm 0.0058$ & $36.7101 \pm 0.4227$ & $9.9373 \pm 0.0554$ & $10.2436 \pm 0.0484$ \\
GaLore-AW32    & $1512.3574 \pm 0.0000$ & $2.2214 \pm 0.0064$ & $25.8785 \pm 0.2514$ & $8.0646 \pm 0.0247$ & $9.0718 \pm 0.0522$ \\
SPAM           & $1512.3574 \pm 0.0000$ & $2.1819 \pm 0.0024$ & $38.7953 \pm 1.3396$ & $7.5490 \pm 0.0787$ & $8.9030 \pm 0.0298$ \\
\end{longtable}
\end{center}


\section{Full Results for Memory-Budget Sweep}
\label{app:budget-sweep-full}

\begin{center}
\scriptsize
\setlength{\tabcolsep}{2.0pt}
\setlength{\LTpre}{0pt}
\setlength{\LTpost}{0pt}
\renewcommand{\arraystretch}{0.98}

\begin{longtable}{@{}cccccc@{}}
\caption{Detailed memory-budget sweep results (mean $\pm$ std). State denotes optimizer-state memory in MB and Time is in seconds. Train/Test performance reports accuracy (\%) for ViT and PPL for GPT-2; it is not applicable (--) for diffusion.}
\label{tab:budget_sweep_full}\\
\toprule
\textbf{Method} & \textbf{State} & \textbf{Loss} & \textbf{Tr. Perf.} & \textbf{Te. Perf.} & \textbf{Time} \\
\midrule
\endfirsthead

\toprule
\textbf{Method} & \textbf{State} & \textbf{Loss} & \textbf{Tr. Perf.} & \textbf{Te. Perf.} & \textbf{Time} \\
\midrule
\endhead

\midrule
\multicolumn{6}{c}{\textit{Continued on next page.}} \\
\endfoot

\bottomrule
\endlastfoot

\multicolumn{6}{c}{\textbf{vit} \textit{(Metric: Accuracy (\%))}} \\
\midrule
Adam-mini        & $41.9064 \pm 0.0000$ & $1.2985 \pm 0.0048$ & $53.2127 \pm 0.3010$ & $52.6433 \pm 1.0179$ & $86.0267 \pm 0.3365$ \\
AdamW16          & $41.8858 \pm 0.0000$ & $1.0916 \pm 0.0114$ & $60.9453 \pm 0.4249$ & $60.7533 \pm 1.0316$ & $86.2380 \pm 1.4323$ \\
AdamW8           & $21.4473 \pm 0.0000$ & $1.1422 \pm 0.0126$ & $59.0313 \pm 0.3880$ & $57.8900 \pm 0.9403$ & $83.3457 \pm 1.2652$ \\
Heu-1            & $20.6921 \pm 1.8592$ & $1.1822 \pm 0.0326$ & $56.0627 \pm 1.2299$ & $56.4200 \pm 0.5323$ & $100.1256 \pm 31.8929$ \\
Heu-2            & $18.3014 \pm 0.0000$ & $1.1830 \pm 0.0057$ & $56.1820 \pm 0.3411$ & $57.2633 \pm 0.6478$ & $86.3452 \pm 0.7325$ \\
GaLore-AF        & $0.3988 \pm 0.0000$  & $1.2719 \pm 0.0071$ & $54.2260 \pm 0.3260$ & $55.1033 \pm 1.3251$ & $90.8836 \pm 0.6302$ \\
GaLore-AW32      & $83.5108 \pm 0.0000$ & $1.1416 \pm 0.0065$ & $59.0827 \pm 0.1462$ & $58.8433 \pm 0.5871$ & $81.8994 \pm 5.1845$ \\
BAOC$_{0.3}$     & $12.4752 \pm 0.0000$ & $1.2497 \pm 0.0134$ & $55.0827 \pm 0.5791$ & $57.3633 \pm 0.4392$ & $83.6926 \pm 0.4624$ \\
BAOC$_{0.4}$     & $16.2369 \pm 0.0000$ & $1.1983 \pm 0.0057$ & $56.9887 \pm 0.2994$ & $57.1533 \pm 1.7723$ & $153.9700 \pm 123.3732$ \\
BAOC$_{0.5}$     & $20.2623 \pm 0.0000$ & $1.1401 \pm 0.0106$ & $59.0427 \pm 0.3119$ & $57.7991 \pm 0.4784$ & $81.3693 \pm 4.7401$ \\
BAOC$_{0.6}$     & $24.7618 \pm 0.0000$ & $1.1318 \pm 0.0227$ & $59.3587 \pm 0.8076$ & $58.4167 \pm 2.3859$ & $151.5932 \pm 116.0048$ \\
BAOC$_{0.7}$     & $28.6280 \pm 0.0000$ & $1.1234 \pm 0.0038$ & $59.8193 \pm 0.2299$ & $59.5133 \pm 1.2051$ & $85.9355 \pm 3.8425$ \\
BAOC$_{0.8}$     & $33.0675 \pm 0.0000$ & $1.1163 \pm 0.0091$ & $60.0680 \pm 0.4291$ & $59.7900 \pm 1.4845$ & $135.7561 \pm 92.4524$ \\
BAOC$_{1.0}$     & $41.8858 \pm 0.0000$ & $1.0901 \pm 0.0190$ & $60.9807 \pm 0.6461$ & $60.8700 \pm 0.3035$ & $86.2148 \pm 2.5320$ \\
BAOC$_{1.2}$     & $49.7608 \pm 0.0000$ & $1.1067 \pm 0.0014$ & $60.4667 \pm 0.1301$ & $60.3367 \pm 1.0373$ & $83.0399 \pm 1.9763$ \\
SPAM             & $83.5108 \pm 0.0000$ & $1.2751 \pm 0.0439$ & $54.0853 \pm 1.6791$ & $52.2067 \pm 1.3518$ & $89.6586 \pm 0.2241$ \\
\addlinespace

\multicolumn{6}{c}{\textbf{diffusion} \textit{(Metric: Loss)}} \\
\midrule
Adam-mini        & $3280.1309 \pm 0.0000$ & $0.2252 \pm 0.0004$ & -- & -- & $128.7334 \pm 0.6368$ \\
AdamW16          & $3280.1685 \pm 0.0000$ & $0.2251 \pm 0.0004$ & -- & -- & $130.5123 \pm 0.5839$ \\
AdamW8           & $1667.6203 \pm 0.0000$ & $0.2358 \pm 0.0017$ & -- & -- & $125.4920 \pm 0.2489$ \\
Heu-1            & $1769.0375 \pm 0.0000$ & $0.2305 \pm 0.0017$ & -- & -- & $129.4730 \pm 0.0861$ \\
Heu-2            & $1551.8149 \pm 0.0000$ & $0.2317 \pm 0.0004$ & -- & -- & $125.8407 \pm 0.1301$ \\
GaLore-AF        & $1745.5663 \pm 0.0000$ & $0.2279 \pm 0.0019$ & -- & -- & $146.8112 \pm 1.4582$ \\
GaLore-AW32      & $6557.6197 \pm 0.0000$ & $0.2251 \pm 0.0010$ & -- & -- & $128.5222 \pm 0.2804$ \\
BAOC$_{0.3}$     & $982.0572 \pm 0.0000$  & $0.2397 \pm 0.0008$ & -- & -- & $125.0603 \pm 0.1528$ \\
BAOC$_{0.5}$     & $1637.8564 \pm 0.0000$ & $0.2329 \pm 0.0019$ & -- & -- & $112.9059 \pm 21.4441$ \\
BAOC$_{0.55}$    & $1786.2088 \pm 0.0000$ & $0.2256 \pm 0.0016$ & -- & -- & $114.4900 \pm 22.4295$ \\
BAOC$_{0.6}$     & $1948.5093 \pm 0.0000$ & $0.2257 \pm 0.0012$ & -- & -- & $112.3163 \pm 23.9905$ \\
BAOC$_{0.7}$     & $2294.8428 \pm 0.0000$ & $0.2259 \pm 0.0014$ & -- & -- & $113.7876 \pm 20.8613$ \\
BAOC$_{0.8}$     & $2622.3323 \pm 0.0000$ & $0.2260 \pm 0.0004$ & -- & -- & $112.5385 \pm 24.1660$ \\
BAOC$_{1.0}$     & $3275.7522 \pm 0.0000$ & $0.2241 \pm 0.0004$ & -- & -- & $112.9285 \pm 23.6251$ \\
BAOC$_{1.2}$     & $3931.2718 \pm 0.0000$ & $0.2249 \pm 0.0019$ & -- & -- & $113.9473 \pm 21.6529$ \\
SPAM             & $6557.6197 \pm 0.0000$ & $0.2272 \pm 0.0017$ & -- & -- & $105.4272 \pm 1.4407$ \\
\addlinespace

\multicolumn{6}{c}{\textbf{gpt2\_small\_alpaca} \textit{(Metric: Perplexity)}} \\
\midrule
Adam-mini        & $475.2570 \pm 0.0000$ & $2.0322 \pm 0.0026$ & $6.4824 \pm 0.0236$ & $7.6449 \pm 0.0105$ & $196.3966 \pm 2.4516$ \\
AdamW16          & $475.1514 \pm 0.0000$ & $1.9699 \pm 0.0004$ & $6.1605 \pm 0.0002$ & $7.4777 \pm 0.0048$ & $199.8222 \pm 1.3254$ \\
AdamW8           & $241.8354 \pm 0.0000$ & $1.9356 \pm 0.0003$ & $5.8695 \pm 0.0034$ & $7.4202 \pm 0.0080$ & $199.5719 \pm 2.1013$ \\
Heu-1            & $249.2868 \pm 0.0000$ & $2.1019 \pm 0.0003$ & $7.0778 \pm 0.0017$ & $7.8163 \pm 0.0022$ & $199.5488 \pm 1.2454$ \\
Heu-2            & $246.6870 \pm 0.0000$ & $2.0661 \pm 0.0003$ & $6.8065 \pm 0.0028$ & $7.7879 \pm 0.0023$ & $199.1805 \pm 0.9018$ \\
GaLore-AF        & $1.2216 \pm 0.0000$   & $2.2012 \pm 0.0004$ & $7.9456 \pm 0.0018$ & $8.3437 \pm 0.0026$ & $214.7760 \pm 0.9798$ \\
GaLore-AW32      & $949.3887 \pm 0.0000$ & $1.9436 \pm 0.0005$ & $5.9097 \pm 0.0026$ & $7.4209 \pm 0.0028$ & $195.9964 \pm 0.8542$ \\
BAOC$_{0.3}$     & $141.6101 \pm 0.0000$ & $2.1810 \pm 0.0014$ & $7.6764 \pm 0.0126$ & $8.1842 \pm 0.0089$ & $203.5953 \pm 7.9204$ \\
BAOC$_{0.4}$     & $189.8923 \pm 0.0000$ & $2.0188 \pm 0.0001$ & $6.4317 \pm 0.0084$ & $7.5776 \pm 0.0118$ & $201.0469 \pm 2.5814$ \\
BAOC$_{0.5}$     & $234.9379 \pm 0.0000$ & $1.9351 \pm 0.0005$ & $5.8345 \pm 0.0061$ & $7.4307 \pm 0.0062$ & $199.2146 \pm 2.0151$ \\
BAOC$_{0.6}$     & $281.6791 \pm 0.0000$ & $1.9330 \pm 0.0002$ & $5.8336 \pm 0.0034$ & $7.4287 \pm 0.0081$ & $201.1650 \pm 1.0886$ \\
BAOC$_{0.7}$     & $321.5229 \pm 0.0000$ & $1.9387 \pm 0.0002$ & $5.8867 \pm 0.0031$ & $7.4350 \pm 0.0054$ & $202.7028 \pm 3.5449$ \\
BAOC$_{0.8}$     & $374.6479 \pm 0.0000$ & $1.9495 \pm 0.0001$ & $5.9754 \pm 0.0067$ & $7.4500 \pm 0.0102$ & $205.5998 \pm 7.4717$ \\
BAOC$_{1.0}$     & $475.1514 \pm 0.0000$ & $1.9697 \pm 0.0002$ & $6.1619 \pm 0.0032$ & $7.4817 \pm 0.0033$ & $201.8412 \pm 2.3359$ \\
BAOC$_{1.2}$     & $556.1514 \pm 0.0000$ & $1.9647 \pm 0.0003$ & $6.1066 \pm 0.0047$ & $7.4655 \pm 0.0076$ & $200.5310 \pm 3.0332$ \\
SPAM             & $949.3887 \pm 0.0000$ & $1.8908 \pm 0.0004$ & $5.4494 \pm 0.0045$ & $7.4163 \pm 0.0125$ & $365.1455 \pm 251.9334$ \\
\addlinespace

\multicolumn{6}{c}{\textbf{gpt2\_small\_gsm8k} \textit{(Metric: Perplexity)}} \\
\midrule
Adam-mini        & $475.2570 \pm 0.0000$ & $1.7656 \pm 0.0168$ & $5.0944 \pm 0.0880$ & $5.3255 \pm 0.0728$ & $30.8223 \pm 0.0937$ \\
AdamW16          & $475.1514 \pm 0.0000$ & $1.5539 \pm 0.0005$ & $4.1222 \pm 0.0016$ & $4.6970 \pm 0.0047$ & $31.7984 \pm 0.1447$ \\
AdamW8           & $241.8354 \pm 0.0000$ & $1.5367 \pm 0.0003$ & $4.0934 \pm 0.0030$ & $4.6912 \pm 0.0023$ & $31.3584 \pm 0.2472$ \\
Heu-1            & $249.2868 \pm 0.0000$ & $1.6980 \pm 0.0010$ & $4.7568 \pm 0.0016$ & $5.0201 \pm 0.0012$ & $31.2432 \pm 0.0651$ \\
Heu-2            & $246.6870 \pm 0.0000$ & $1.7087 \pm 0.0008$ & $4.8158 \pm 0.0028$ & $5.2329 \pm 0.0030$ & $31.7842 \pm 0.1475$ \\
GaLore-AF        & $1.2216 \pm 0.0000$   & $1.8629 \pm 0.0005$ & $5.6228 \pm 0.0005$ & $5.7685 \pm 0.0007$ & $33.7546 \pm 0.3748$ \\
GaLore-AW32      & $949.3887 \pm 0.0000$ & $1.5582 \pm 0.0003$ & $4.1314 \pm 0.0056$ & $4.6956 \pm 0.0084$ & $30.9713 \pm 0.2742$ \\
BAOC$_{0.3}$     & $141.6101 \pm 0.0000$ & $1.8189 \pm 0.0005$ & $5.2427 \pm 0.0055$ & $5.4252 \pm 0.0054$ & $31.3703 \pm 0.6086$ \\
BAOC$_{0.4}$     & $189.8923 \pm 0.0000$ & $1.6079 \pm 0.0003$ & $4.3424 \pm 0.0015$ & $4.7354 \pm 0.0031$ & $31.6515 \pm 0.5216$ \\
BAOC$_{0.5}$     & $234.9379 \pm 0.0000$ & $1.5381 \pm 0.0003$ & $4.0870 \pm 0.0015$ & $4.6882 \pm 0.0047$ & $31.4935 \pm 0.1790$ \\
BAOC$_{0.6}$     & $281.6791 \pm 0.0000$ & $1.5405 \pm 0.0003$ & $4.0604 \pm 0.0050$ & $4.6689 \pm 0.0082$ & $31.8960 \pm 0.9961$ \\
BAOC$_{0.7}$     & $321.5229 \pm 0.0000$ & $1.5437 \pm 0.0010$ & $4.0784 \pm 0.0009$ & $4.6834 \pm 0.0047$ & $35.1542 \pm 6.7310$ \\
BAOC$_{0.8}$     & $374.6479 \pm 0.0000$ & $1.5484 \pm 0.0006$ & $4.1046 \pm 0.0034$ & $4.6980 \pm 0.0037$ & $32.1326 \pm 0.8763$ \\
BAOC$_{1.0}$     & $475.1514 \pm 0.0000$ & $1.5543 \pm 0.0008$ & $4.1276 \pm 0.0021$ & $4.7016 \pm 0.0016$ & $31.9755 \pm 0.7810$ \\
BAOC$_{1.2}$     & $556.1514 \pm 0.0000$ & $1.5520 \pm 0.0006$ & $4.1163 \pm 0.0055$ & $4.6933 \pm 0.0078$ & $32.5971 \pm 1.7064$ \\
SPAM             & $949.3887 \pm 0.0000$ & $1.5661 \pm 0.0007$ & $4.1820 \pm 0.0028$ & $4.7273 \pm 0.0056$ & $33.1966 \pm 0.1087$ \\
\end{longtable}
\end{center}


\section{Additional Experimental Diagnostics}
\label{app:add-exp}

The rows in Table~\ref{tab:exp3_results} report the diagnostic settings whose selected allocation differs from the default BAOC$_{0.5}$ allocation. We also evaluated additional sampling ratios and Agg weights that led to the same selected allocation; these are omitted to avoid redundant rows. In particular, the sampling-ratio sweep covers $s\in\{0.1\%,1\%,2\%,5\%,10\%\}$, and the Agg sweep covers $\gamma\in\{0.1,0.2,\ldots,0.9\}$; only the settings that changed the allocation are listed below.

\begin{center}
\scriptsize
\setlength{\tabcolsep}{2.0pt}
\setlength{\LTpre}{0pt}
\setlength{\LTpost}{0pt}
\renewcommand{\arraystretch}{0.95}

\begin{longtable}{@{}ccccccc@{}}
\caption{Additional diagnostic results (mean $\pm$ std). Only settings that change the selected allocation relative to BAOC$_{0.5}$ are listed. State denotes optimizer-state memory (MB), and Peak denotes peak end-to-end GPU memory footprint (MB). For ViT, Train/Test report accuracy (\%); for language workloads, they report PPL; for diffusion, only loss is applicable.}
\label{tab:exp3_results}\\
\toprule
\textbf{Variant} & \textbf{State} & \textbf{Loss} & \textbf{Train} & \textbf{Test} & \textbf{Time} & \textbf{Peak} \\
\midrule
\endfirsthead

\toprule
\textbf{Variant} & \textbf{State} & \textbf{Loss} & \textbf{Train} & \textbf{Test} & \textbf{Time} & \textbf{Peak} \\
\midrule
\endhead

\midrule
\multicolumn{7}{c}{\textit{Continued on next page.}}\\
\endfoot

\bottomrule
\endlastfoot

\multicolumn{7}{c}{\textbf{ViT} \textit{(Metric: Accuracy (\%))}}\\
\midrule
BAOC$_{0.5}$ &
\makecell[c]{20.2623\\$\pm$0.0000} &
\makecell[c]{1.1401\\$\pm$0.0106} &
\makecell[c]{59.0427\\$\pm$0.3119} &
\makecell[c]{57.7991\\$\pm$0.4784} &
\makecell[c]{81.3693\\$\pm$4.7401} &
\makecell[c]{120.4912\\$\pm$0.2691} \\
BAOC-online &
\makecell[c]{21.9991\\$\pm$0.0000} &
\makecell[c]{1.3616\\$\pm$0.0403} &
\makecell[c]{51.3000\\$\pm$1.4951} &
\makecell[c]{53.8933\\$\pm$1.1957} &
\makecell[c]{65.3324\\$\pm$1.5700} &
\makecell[c]{128.5718\\$\pm$0.4719} \\
\addlinespace

\multicolumn{7}{c}{\textbf{Diffusion (UNet)} \textit{(Metric: Loss)}}\\
\midrule
BAOC$_{0.5}$ &
\makecell[c]{1637.8564\\$\pm$0.0000} &
\makecell[c]{0.2329\\$\pm$0.0019} &
-- & -- &
\makecell[c]{112.9059\\$\pm$21.4441} &
\makecell[c]{12402.4512\\$\pm$0.0180} \\
BAOC$_{0.5}$ ($\gamma=0.9$) &
\makecell[c]{1637.9016\\$\pm$0.0000} &
\makecell[c]{0.2468\\$\pm$0.0019} &
-- & -- &
\makecell[c]{84.5026\\$\pm$0.1730} &
\makecell[c]{12438.7300\\$\pm$75.7273} \\
BAOC-pref &
\makecell[c]{1751.1073\\$\pm$0.0000} &
\makecell[c]{0.2259\\$\pm$0.0009} &
-- & -- &
\makecell[c]{145.7074\\$\pm$1.0211} &
\makecell[c]{12427.6187\\$\pm$77.5449} \\
BAOC-no-SGD &
\makecell[c]{1786.2088\\$\pm$0.0000} &
\makecell[c]{0.2245\\$\pm$0.0017} &
-- & -- &
\makecell[c]{88.5910\\$\pm$1.7018} &
\makecell[c]{10237.5407\\$\pm$44.9734} \\
\addlinespace

\multicolumn{7}{c}{\textbf{GPT2-small (GSM8K)} \textit{(Metric: Perplexity)}}\\
\midrule
BAOC$_{0.5}$ &
\makecell[c]{234.9379\\$\pm$0.0000} &
\makecell[c]{1.5381\\$\pm$0.0003} &
\makecell[c]{4.0870\\$\pm$0.0015} &
\makecell[c]{4.6882\\$\pm$0.0047} &
\makecell[c]{31.4935\\$\pm$0.1790} &
\makecell[c]{1675.8228\\$\pm$46.8145} \\
BAOC$_{0.5}$ ($\gamma=0.9$) &
\makecell[c]{234.9379\\$\pm$0.0000} &
\makecell[c]{1.5513\\$\pm$0.0007} &
\makecell[c]{4.1003\\$\pm$0.0011} &
\makecell[c]{4.6979\\$\pm$0.0019} &
\makecell[c]{31.0567\\$\pm$0.0820} &
\makecell[c]{1703.0807\\$\pm$0.8057} \\
BAOC-online &
\makecell[c]{255.1166\\$\pm$0.0000} &
\makecell[c]{1.5374\\$\pm$0.0269} &
\makecell[c]{4.0183\\$\pm$0.1065} &
\makecell[c]{4.6538\\$\pm$0.0604} &
\makecell[c]{48.6532\\$\pm$0.2420} &
\makecell[c]{1858.1430\\$\pm$1.9824} \\
\addlinespace

\multicolumn{7}{c}{\textbf{T5-base (GSM8K)} \textit{(Metric: Perplexity)}}\\
\midrule
BAOC$_{0.5}$ &
\makecell[c]{377.7700\\$\pm$0.0000} &
\makecell[c]{2.2029\\$\pm$0.0098} &
\makecell[c]{7.8627\\$\pm$0.1020} &
\makecell[c]{8.9859\\$\pm$0.0457} &
\makecell[c]{27.0551\\$\pm$0.2333} &
\makecell[c]{2792.4321\\$\pm$0.0015} \\
BAOC-online &
\makecell[c]{415.6152\\$\pm$0.0000} &
\makecell[c]{2.2090\\$\pm$0.0083} &
\makecell[c]{7.8994\\$\pm$0.0453} &
\makecell[c]{9.0595\\$\pm$0.0750} &
\makecell[c]{23.0398\\$\pm$0.6264} &
\makecell[c]{3014.5227\\$\pm$19.2073} \\
BAOC$_{0.5}$ ($s=1\%$) &
\makecell[c]{378.0718\\$\pm$0.0000} &
\makecell[c]{2.2895\\$\pm$0.0055} &
\makecell[c]{8.8116\\$\pm$0.0661} &
\makecell[c]{9.6051\\$\pm$0.1174} &
\makecell[c]{27.2872\\$\pm$1.0093} &
\makecell[c]{2792.4321\\$\pm$0.0051} \\
BAOC$_{0.5}$ ($s=2\%$) &
\makecell[c]{377.9907\\$\pm$0.0000} &
\makecell[c]{2.2821\\$\pm$0.0054} &
\makecell[c]{8.6982\\$\pm$0.0504} &
\makecell[c]{9.5002\\$\pm$0.0824} &
\makecell[c]{27.3051\\$\pm$0.2655} &
\makecell[c]{2791.4565\\$\pm$1.6898} \\
BAOC$_{0.5}$ ($s=5\%$) &
\makecell[c]{375.0648\\$\pm$0.0000} &
\makecell[c]{2.2895\\$\pm$0.0055} &
\makecell[c]{8.4355\\$\pm$0.0531} &
\makecell[c]{9.4049\\$\pm$0.1504} &
\makecell[c]{26.1808\\$\pm$0.2555} &
\makecell[c]{2636.8529\\$\pm$9.4712} \\
\end{longtable}
\end{center}

\subsection{Anchor perturbation}
\label{app:anchor-perturbation}

We further perturb the two anisotropy anchors used in the normalization step. The results in Table~\ref{tab:anchor_perturbation} use the same evaluation protocol as the corresponding BAOC runs. The values are reported exactly as in the rebuttal-stage additional experiment.

\begin{table}[t]
\centering
\scriptsize
\setlength{\tabcolsep}{3pt}
\caption{Sensitivity to anisotropy-anchor perturbation. For ViT, the metric is test accuracy (\%, higher is better). For GPT2-small/GSM8K, the metric is test perplexity (lower is better). PeakMem is in MB.}
\label{tab:anchor_perturbation}
\begin{tabular}{@{}ccccc@{}}
\toprule
\textbf{Task} & \textbf{Setting} & \textbf{Loss} & \textbf{Metric} & \textbf{PeakMem} \\
\midrule
ViT & BAOC$_{0.5}$ & 1.1401 & Acc: 57.7991 & 120.4912 \\
ViT & $+20\%$ & 1.1427 & Acc: 57.4754 & 117.5808 \\
ViT & $-20\%$ & 1.1469 & Acc: 57.2384 & 121.6471 \\
\midrule
GPT2-small/GSM8K & BAOC$_{0.5}$ & 1.5381 & PPL: 4.6882 & 1675.8228 \\
GPT2-small/GSM8K & $+20\%$ & 1.5736 & PPL: 4.9072 & 1702.3024 \\
GPT2-small/GSM8K & $-10\%$ & 1.5204 & PPL: 4.6693 & 1687.2187 \\
GPT2-small/GSM8K & $-20\%$ & 1.5989 & PPL: 4.7266 & 1748.3895 \\
\bottomrule
\end{tabular}
\end{table}

\subsection{Effect of block partitioning}
\label{app:partition}

BAOC requires a partition of the trainable parameters before computing block-level diagnostics and solving the allocation problem.
The partition should be fine enough to preserve heterogeneous optimizer-risk signals, but not so fine that the diagnostics become noisy or the allocation becomes fragmented.
Overly coarse partitioning may merge components with different gradient behavior, such as attention and FFN modules, thereby averaging anisotropy and direction-stability signals into a less discriminative compromise value.
Overly fine partitioning, such as parameter-tensor-level partitioning, can produce blocks with too few sampled coordinates, making the EMA-based diagnostics sensitive to warmup noise and increasing the number of MILP variables without clear practical benefit.
We therefore use a structure-guided and statistic-refined partitioning procedure.

Let $\Theta$ denote the set of trainable parameters and $|\Theta|$ its total parameter count.
BAOC first scans the model structure to generate candidate units, computes normalized diagnostic vectors from warmup gradient streams, and then forms final blocks by combining structural boundaries, statistic differences, and a minimum block-size constraint.
The procedure is summarized in Algorithm~\ref{alg:block_partition}.

\begin{algorithm}[t]
\caption{BAOC block partitioning}
\label{alg:block_partition}
\begin{algorithmic}[1]
\STATE \textbf{Input:} instantiated model $f_\theta$; warmup gradient streams; diagnostic weights $\{w_k\}$; minimum block-size threshold $n_{\min}$; statistic-difference threshold $\tau$
\STATE \textbf{Output:} parameter block set $\mathcal{B}=\{\mathcal{B}_i\}_{i=1}^N$

\STATE \textbf{Structural scan:}
\STATE Traverse the forward structure of $f_\theta$ and form initial structural units $\mathcal{U}=\{U_j\}_{j=1}^m$
\STATE Record the parameter count $n_j=|U_j|$ for each unit $U_j$

\STATE \textbf{Warmup diagnostics:}
\FOR{each unit $U_j\in\mathcal{U}$}
    \STATE Collect sampled gradient streams during warmup
    \STATE Compute and normalize the diagnostic vector $z_j=(z_{j,1},\ldots,z_{j,K})$
\ENDFOR

\STATE \textbf{Statistic difference:}
\FOR{each pair of candidate units $(U_a,U_b)$}
    \STATE Compute the weighted maximum component difference
    \[
    \Delta(U_a,U_b)=\max_{k=1,\ldots,K} w_k\,|z_{a,k}-z_{b,k}|
    \]
\ENDFOR

\STATE \textbf{Candidate boundary selection:}
\STATE Generate candidate boundaries from model structural boundaries and pairs with large statistic difference
\STATE In our implementation, $\tau$ is set to the upper quartile of all pairwise values $\Delta(U_a,U_b)$ among candidate units, and $w_k=1$ by default

\STATE \textbf{Minimum-size filtering:}
\FOR{each candidate unit or temporary block $U$}
    \IF{$|U|<n_{\min}$}
        \STATE Merge $U$ into the statistically closest neighboring candidate block that yields a more appropriate size
    \ENDIF
\ENDFOR

\STATE \textbf{Block formation:}
\STATE Initialize $\mathcal{B}\leftarrow \emptyset$
\FOR{adjacent candidate units or temporary blocks $(U_a,U_b)$}
    \IF{$\Delta(U_a,U_b)>\tau$ and both units satisfy the minimum-size constraint}
        \STATE Keep $U_a$ and $U_b$ as separate blocks
    \ELSE
        \STATE Merge them into the same block
    \ENDIF
\ENDFOR

\STATE \textbf{Post-processing:}
\STATE Split blocks with clearly separated internal statistics
\STATE Re-merge blocks whose parameter count remains below $n_{\min}$
\STATE Return the final block set $\mathcal{B}$
\end{algorithmic}
\end{algorithm}

The minimum-size threshold prevents small blocks from producing unstable diagnostics under sparse coordinate sampling.
In our implementation, we parameterize it as
\[
n_{\min}=\alpha|\Theta|,
\]
where $\alpha$ is a small implementation-level ratio and $|\Theta|$ denotes the total number of trainable parameters.
This threshold is not used to impose semantic partitions; it only prevents statistically unreliable or overly fragmented blocks.

For Transformer-based language models, the structural scan first generates candidate units according to the forward structure of each Transformer block.
Attention, FFN, embedding, LM head, and normalization modules provide initial candidate boundaries, but they do not force the final partition.
BAOC makes partitioning decisions layer by layer and position by position.
Within each layer, a candidate unit is retained as an independent block only if it satisfies the minimum-size constraint and exhibits sufficiently different diagnostics from neighboring units according to $\Delta(\cdot,\cdot)$.
Thus, structural priors determine where candidate boundaries are checked, while warmup statistics and size constraints determine the final block set.

Table~\ref{tab:partition_granularity} compares the default statistic-refined BAOC partitioning with a coarser partitioning strategy on language workloads.
The comparison uses final test PPL as the task metric.
The coarser strategy merges structurally adjacent units more aggressively, which tends to average heterogeneous gradient statistics within a larger block and weakens the discriminative power of the block-level risk signals.
The results show that the default partitioning consistently improves over the coarse partition.

\begin{table}[t]
\centering
\scriptsize
\setlength{\tabcolsep}{3pt}
\caption{Effect of block partitioning granularity. Lower PPL is better.}
\label{tab:partition_granularity}
\begin{tabular}{@{}ccc@{}}
\toprule
\textbf{Task} & \textbf{BAOC$_{0.5}$ PPL} & \textbf{BAOC$_{0.5}$ coarse PPL} \\
\midrule
GPT2-small/Alpaca & 7.3907 & 7.6156 \\
GPT2-small/GSM8K & 4.5980 & 4.7928 \\
T5-base/Alpaca & 8.2520 & 8.2813 \\
T5-base/GSM8K & 9.2202 & 9.4216 \\
\bottomrule
\end{tabular}
\end{table}

\subsection{Online reallocation with state inheritance}
\label{app:online-inheritance}

Table~\ref{tab:online_inheritance} reports the additional inheritance study for online reallocation. The no-inheritance setting restarts optimizer states after each reallocation, whereas the inheritance setting migrates compatible optimizer states when possible. For ViT, the task metric is test accuracy; for GPT2-small/GSM8K, the task metric is test PPL.

\begin{table}[t]
\centering
\scriptsize
\setlength{\tabcolsep}{3pt}
\caption{Online reallocation with and without optimizer-state inheritance. PeakMem is in MB. Higher accuracy is better and lower PPL is better.}
\label{tab:online_inheritance}
\begin{tabular}{@{}ccccc@{}}
\toprule
\textbf{Method} & \textbf{Loss} & \textbf{Test Accuracy} & \textbf{Test PPL} & \textbf{PeakMem} \\
\midrule
\multicolumn{5}{c}{\textbf{ViT}} \\
\midrule
BAOC$_{0.5}$ & $1.1401 \pm 0.0106$ & $57.7991 \pm 0.4784$ & -- & $120.4912 \pm 0.2691$ \\
BAOC-online & $1.3616 \pm 0.0403$ & $53.8933 \pm 1.1957$ & -- & $128.5718 \pm 0.4719$ \\
BAOC-online+inherit & $1.1209 \pm 0.0155$ & $57.8288 \pm 0.4542$ & -- & $126.1732 \pm 0.3505$ \\
\midrule
\multicolumn{5}{c}{\textbf{GPT2-small/GSM8K}} \\
\midrule
BAOC$_{0.5}$ & $1.5381 \pm 0.0003$ & -- & $4.6882 \pm 0.0047$ & $1675.8228 \pm 46.8145$ \\
BAOC-online & $1.5374 \pm 0.0269$ & -- & $4.6538 \pm 0.0604$ & $1858.1430 \pm 1.9824$ \\
BAOC-online+inherit & $1.5309 \pm 0.0046$ & -- & $4.6527 \pm 0.0425$ & $1721.8571 \pm 48.6652$ \\
\bottomrule
\end{tabular}
\end{table}

\end{document}